\documentclass[a4paper,conference]{IEEEtran}
\renewcommand{\baselinestretch}{0.97} 

\ifCLASSINFOpdf
\else
\fi

\usepackage[colorlinks]{hyperref}
\usepackage{graphicx}
\usepackage{caption}

\usepackage{atbegshi,picture}
\usepackage{lipsum}

\AtBeginShipout{\AtBeginShipoutUpperLeft{%
  \put(\dimexpr\paperwidth-13.7cm\relax,-1.2cm){\makebox[0pt][r]{Accepted as Demo paper at ICPR 2020}}%
}}

\begin{document}
%
\title{From Tinkering to Engineering: \\Measurements in Tensorflow Playground }

\author{
\IEEEauthorblockN{Axel Harstad}
\IEEEauthorblockA{
University of California, Berkeley \\ and NTNU \\
axeloh@berkeley.edu}
\and
\IEEEauthorblockN{Henrik Hoeiness}
\IEEEauthorblockA{
University of California, Berkeley \\ and NTNU \\
henrhoi@berkeley.edu}
\and
\IEEEauthorblockN{Gerald Friedland}
\IEEEauthorblockA{University of California, Berkeley\\
fractor@eecs.berkeley.edu}
}

\maketitle

\begin{abstract}
In this article, we present an extension of the Tensorflow Playground~\cite{tf_playground_paper}, called Tensorflow Meter (short TFMeter). TFMeter is an interactive neural network architecting tool that allows the visual creation of different architecture of neural networks. In addition to its ancestor, the playground, our tool shows information-theoretic measurements while constructing, training, and testing the network. As a result, each change results in a change in at least one of the measurements, providing for a better engineering intuition of what different architectures are able to learn. The measurements are derived from various places in the literature, including recent work by \cite{friedland_paper1}. In this paper, we describe our web application that is available online\cite{tfmeter} and argue that in the same way that the original Playground is meant to build an intuition about neural networks, our extension educates users on available measurements, which we hope will ultimately improve experimental design and reproducibility in the field.
\end{abstract}

\section{Introduction}
When performing neural network experiments many details need to be addressed, such as the number of neurons, network topology, activation function, and secondary parameters like learning rate or regularization. Today, the most common way to find these parameters seems to be through a ``guess and check” approach. While this can work fine anecdotally, it is certainly not ideal, as parameter search can take exponential time, burns billions of computing cycles, may lead to oversized networks where energy is wasted during both training and testing, and, if the experiments do not work, it is back to the drawing board. As a result, there is currently a movement to improve the reproducibility of machine learning experiments through engineering measurements. 
 
\begin{figure*}
\centering
\captionsetup{width=0.8\textwidth}
\includegraphics[width=0.8\textwidth]{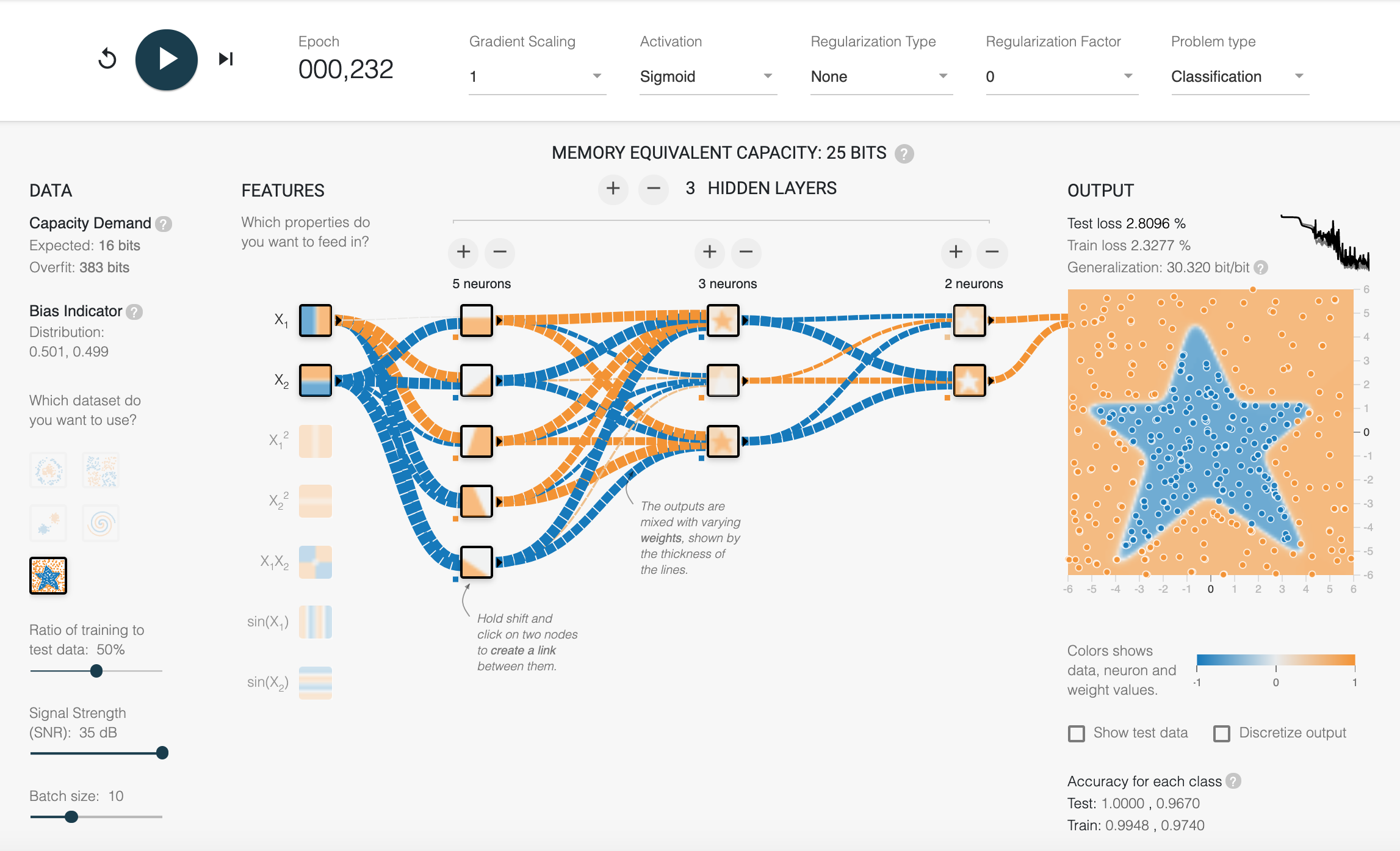}
\caption{TFMeter \protect\cite{tfmeter} in action. This network is classifying data from a dataset uploaded by the user.  Neurons biases and weights are visualized with heatmaps and boldness of connections.The top-right corner shows a graph plotting the loss over time along with the network's generalization ratio. The network's Memory Equivalent Capacity is found in the top-middle section, and the dataset's expected capacity demand in the top-left corner, along with a bias indicator. Classification accuracies for each class are shown bottom-right. \label{cap:bigpicture}}
\label{tfmeter}
\end{figure*}

In accordance with this idea, we present our extension to the Tensorflow Playground \cite{tf_playground_paper,tf_playground_web}, which we call TFMeter (see Figure~\ref{cap:bigpicture}. In our extension, all functionality of the original Tensorflow Playground is preserved but we introduce information-theoretic measurements, such as Memory Equivalent Capacity and bitwise generalization. These measures were presented in various places in the literature, such as \cite{mackay,cover,friedland_paper1,friedland_paper2}. In addition, TFMeter features the ability to manually remove network links, and add direct connections (to simulate a Residual Network). Furthermore, it is possible to run the architected network on one's own dataset. We also indicate bias by measuring class balance and comparing it to the accuracy for each class. This gives the user a flexible platform where one can obtain an intuition for neural networks and deep learning. 

\section{Measurements}
\begin{figure}
\centering 
\includegraphics[width=0.8\linewidth]{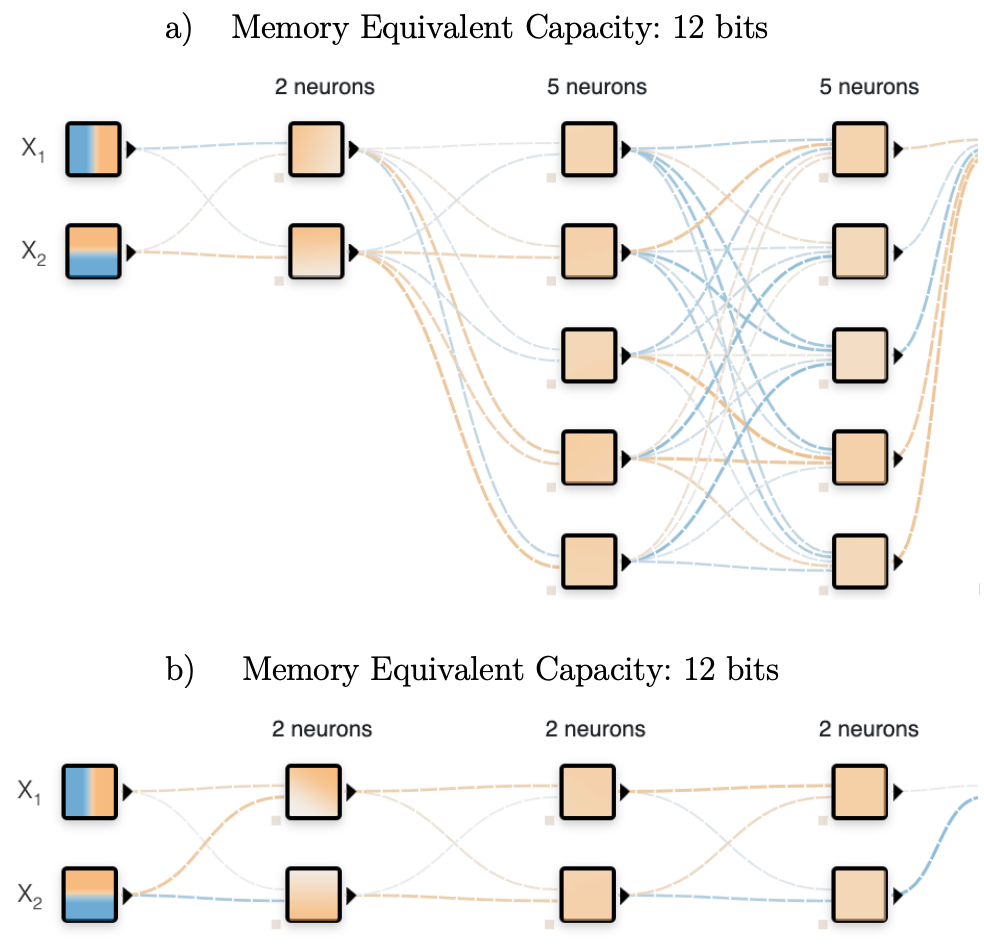}
\caption{Two networks illustrating Memory Equivalent Capacity: In both networks, the second layer is dependent on the first hidden layer. The greater number of neurons in a) makes no difference to the overall capacity of the network, so both networks end up with equal capacity.}
\label{networks_mec_compare}
\end{figure}

The capacity of a Neural Network is defined as the number of functions a network can implement. TFMeter \cite{tfmeter}, just like its ancestor, only supports binary classification. This allows us to adopt the definition of capacity presented in \cite{cover,mackay}, where individual neurons are treated as electrical elements and thus a network's memorization capacity (number of points it can overfit) can be calculated using circuit rules~\cite{friedland_paper1}. For a given neural network architecture, we use these engineering rules to compute the memorization capacity of a network or Memory Equivalent Capacity (MEC). In TFMeter, the MEC is shown just above the corresponding network, see \autoref{tfmeter}.

\begin{figure}
\centering 
\includegraphics[width=0.81\linewidth]{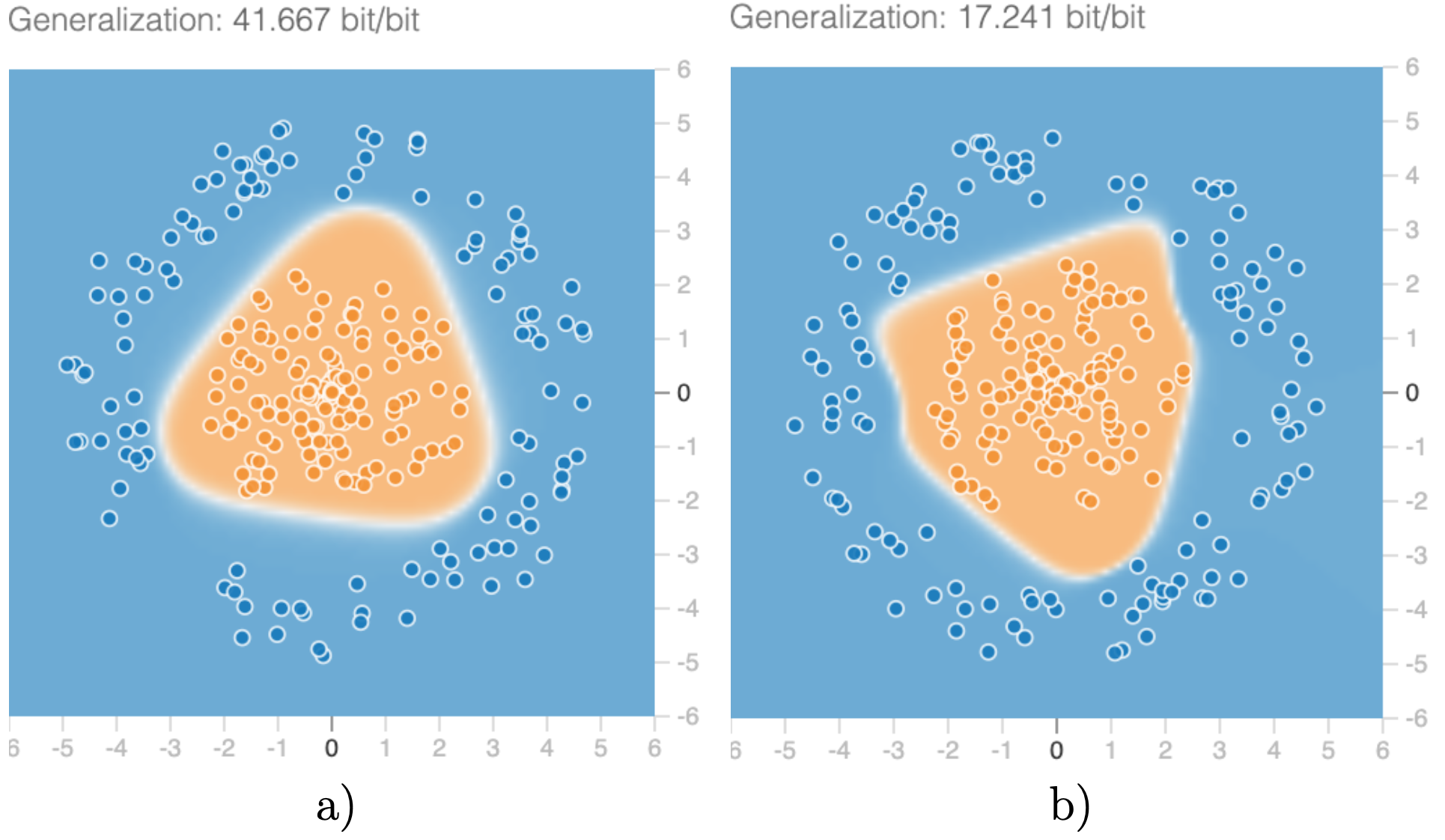}
\caption{Classifications of two networks. The  Memory Equivalent Capacity of network a) is set to equal the dataset's expected capacity demand and b) is above it. Network a) yields a generalization ratio of $G=41.67$, while b) yields $G=17.24$. Both networks reach $100\,\%$ accuracy.}
\label{ComparingMEC}
\end{figure}

MEC allows us to compare different network architectures independent of a task (dataset and labels). For example, consider the two networks in \autoref{networks_mec_compare}. The first network has more neurons in the subsequent layers than the second network, but their MEC is the same. This indicates that the second network is more efficient than the first, as using the first network would result in wasting memory and computation resources. Furthermore, as the second network has less parameters, it should have a better chance to avoid overfitting \cite{friedland_paper1}. 


We have seen that MEC allows the comparison of different architectures independent of the task. However, sizing a network properly to a task requires an estimate of the required capacity. \cite{friedland_paper2} proposes a heuristic method to estimate the required neural network capacity for any given dataset and binary labeling. Experimental results for various datasets show that it works well for estimating the expected required capacity. Therefore, we use this heuristic method to estimate the expected capacity demand for a given dataset. If features are selected, TFMeter takes that into account. The expected capacity demand is shown in the top-left corner, see  \autoref{tfmeter}.

The expected capacity requirement of a dataset is useful because it provides an estimate for how large the network needs to be. For example, in \autoref{ComparingMEC}, we used two different networks for the same binary classification task. The expected required capacity for this dataset was 12 bits. The first network had a MEC equal to the expected required capacity, while the second network had a MEC of 35 bits. Both networks reached $100\,\%$ test accuracy. That is, the larger network is oversized. This relation between network capacity and performance is quantifiable and is called generalization. 

TFMeter \cite{tfmeter} uses the measurement of generalization defined in \cite{friedland_paper1}, which is $G = \frac{\#\ correctly\ predicted\ instances}{Memory\ Equivalent\ Capacity}$. A machine learner with generalization $G \leq 1$ implies no generalization, while $G > 1$ implies successful generalization.

The machine learner with the lowest capacity and highest accuracy is the one that uses the representation function(s) most effectively. Therefore, it has the lowest chance of failing when applied to unseen data. This is consistent with Occam’s razor \cite{occam}, which dictates that for two competing correct hypotheses one should follow the one with the least assumptions.


\end{document}